\documentclass{article}

\usepackage[final]{nips_2017}

\usepackage[utf8]{inputenc} 
\usepackage[T1]{fontenc}    
\usepackage{hyperref}       
\usepackage{url}            
\usepackage{booktabs}       
\usepackage{amsfonts}       
\usepackage{nicefrac}       
\usepackage{microtype}      
\usepackage{comment}
\usepackage{chngpage}

\title{AWE-CM Vectors: Augmenting Word Embeddings with a Clinical Metathesaurus}

\author{
  Willie Boag \\
  CSAIL\\
  MIT\\
  Cambridge, MA \\
  \texttt{wboag@mit.edu} \\
  \And
  Hassan Kané \\
  CSAIL\\
  MIT\\
  Cambridge, MA \\
  \texttt{mhkane@mit.edu} \\
}

\begin{document}

\maketitle

\begin{abstract}

In recent years, word embeddings have been surprisingly effective at capturing intuitive characteristics of the words they represent. These vectors achieve the best results when training corpora are extremely large, sometimes billions of words. Clinical natural language processing datasets, however, tend to be much smaller. Even the largest publicly-available dataset of medical notes is three orders of magnitude smaller than the dataset of the oft-used ``Google News'' word vectors. In order to make up for limited training data sizes, we encode expert domain knowledge into our embeddings. Building on a previous extension of word2vec, we show that generalizing the notion of a word's ``context'' to include arbitrary features creates an avenue for encoding domain knowledge into word embeddings. We show that the word vectors produced by this method outperform their text-only counterparts across the board in correlation with clinical experts.
    
\end{abstract}


\section{Introduction}

Words are the basic building block for concepts, sentences, paragraphs, stories, and further higher-level representation. Whether it's understanding the prose in a clinician's notes or searching keywords in PubMed's metadata, the most effective tools always begin with a useful representation of individual words. For the past few years, distributed word embeddings have become the de facto representation, largely because of how well they seem to reflect intuitive relationships between words. However, the quality of these vectors are highly dependent on the data that they were trained on. The popular and widely used pretrained word embeddings released by Google were built on a news corpus, and the resulting vectors perform terribly for clinical language. Unfortunately even though models do best with large training sets, the largest publicly available dataset of clinical notes is still 3 orders of magnitude smaller than the Google News corpus.

Typically, the less data available, the more heavily one must rely on a priori domain knowledge. This is certainly true in the clinical domain, where many standard classifications and nomenclatures have been built by expert physicians, such as SNOMED, ICD, RxNorm, MeSH, UMLS, and more. The Unified Medical Language System (UMLS) is of particular interest, because it serves as a metathesaurus to link concepts across other existing ontologies when various terms exist to describe a single concept. For instance, the phrases \textit{heart attack}, \textit{cardiac infarction}, \textit{myocardial necrosis}, and many more all get mapped to the unique concept identify or ``CUI'' C0027051.

In this work, we propose a domain-knowledge-injection method and demonstrate that it improves the agreement between word embeddings and clinical expert judgment. These embeddings are built to harness the strengths of both word2vec (prose co-occurrence) and domain expertise (UMLS metathesaurus). Our code is publicly available for download at \url{https://github.com/wboag/awecm}.

\section{Related Work}

\subsection{Word Embeddings}

Historically, words have been treated as atomic symbols using sparse multi-hot feature representations, but such methods fail to detect any notion of similarity between words with similar meanings \citep{socher-thesis}. The advent of distributional representations derived from the co-occurrence and usage statistics signified an important step forward for more semantically-focused representations. In recent years, so-called ``distributed'' representations of words have gained popularity because of their effectiveness at embedding words into a low-dimensional vector space \citep{bengio2003nlm,word2vec2013,glove2014}. These methods take a ``predict-based'' approach to quantifying the co-occurrence relationships between words \citep{baroni2014}, but in 2014--2015 these methods were shown to be more-or-less equivalent to the traditional count-based ``distributional'' representations  \citep{sppmi2014,lessonslearned2015}.

Perhaps more impressive than word2vec's inital results was the vast influx of followup work it inspired in shallow representation learning. Many have focused on making embedding spaces more interprettable so that we can better understand what is being learned \citep{interpret1,interpret2,interpret3}. Others have aimed to models sense-aware vectors \citep{sense:huang2012,sense:mccallum2014,sense:jurafsky2015} and study biases of our learned representations \citep{bias:nips2016}. Further still, researchers have generalized the learning process, such as Levy and Goldberg generalizing a word's ``context'' to include not just words positionally adjacent in a sentence but to structured syntactic ``contexts'' for words to appear in \citep{lg:dependency2014}.

\subsection{Clinical Embeddings}

There have been multiple attempts to encode clinical concepts in vector space models. Nearly a decade ago, efforts from UMinnesota created a dataset of clinician judgments of concepts to assess both path-based ontology similarity metrics and distributional vector space metrics \citep{uminn:data}. In followup work, they explored the effect that first-order vs second-order context definitions had on distributional word vectors \citep{uminn:similarity}.

More recently, Choi et. al. learned dense, low-dimensional embeddings, not of words but of concepts \citep{sontag:cuis}. Three different vector spaces were learned: CUIs from UMLS-processed clinical journal abstracts, medical claims from billing codes, and CUIs from UML-processed clinical notes, and all embeddings are available publicly.\footnote{https://github.com/clinicalml/embeddings} Alternatively, other work has strongly leveraged ontologies by retrofitting word vectors that were built from MeSH Terms in order to better correlate with human judgment \citep{retrofitting2016}.

\section{Data}
\label{sec:data}

This work uses data from the publicly-available Medical Information Mart for Intensive Care (MIMIC-III) database \citep{johnson2016mimiciii}. MIMIC-III v1.4 contains de-identified EHR data from over 58,000 hospital admissions for nearly 38,600 adult patients. The data was collected from Beth Israel Deaconess Medical Center from 2001--2012, and consists of 2 million notes totaling 500 million tokens. Each note was preprocessed by removing PHI tags, collapsing all-caps phrases into a single token, reducing common age regular expressions to per-decade age tokens, removing all non-alphanumeric characters, and normalizing all non-age numbers to zero.


We evaluate word vectors using the MayoSRS dataset containing three sets of clinician judgments, created by the University of Minnesota.\footnote{http://rxinformatics.umn.edu/SemanticRelatednessResources.html} There is a large MayoSRS list of coder similarity scores as well as two MiniMayoSRS lists, one created by physicians and one by coders. These various lists capture a more robust set of expert judgments. Each set is a list of concept-concept pairs and an expert-assigned score denoting either the similarity or relatedness between those two concepts. Vector quality is assessed by computing the correlation between expert judgments and the cosine similarities of the embeddings of the clinical phrases.\footnote{When a phrase is multiple words, the phrase embedding is computed as the average of its word embeddings.}

\section{Methods}

The goal of this work is to incorporate domain knowledge into word embeddings. We explore an extension of word2vec that allows training on arbitrary feature ``contexts'', and show that adding UMLS-derived features improves performance. We train 3 word2vec-based models on a corpus of 505 million tokens from MIMIC III. For all experiments, hyperparameters are held constant for fair evaluation: 300 dimensions, a dynamic context window of 8 tokens, a subsampling rate of 1e-4, a min-count of 5, 5 iterations, 8 negative samples, the skip-gram architecture, and an alpha of 0.025.

\subsection{word2vec}

The original word2vec tool was based on a simple idea: similar words appear in similar situations. Even before distributed representations, count-based ``distributional representations'' revolved around the concept of knowing a word ``by the company it keeps.'' More concretely, the word ``car'' tends to appear next to words such as \textit{drive}, \textit{gas}, and \textit{fast}; the word ``truck'' tends to appear next to \textit{drive}, \textit{gas}, and \textit{strong}. The shared neighbors for ``car'' and ``truck'' demonstrate the strong overlap between those two concepts.

The insight for word2vec was to frame the distributional hypothesis in a predictive, rather than count-based approach. Instead of using statistics to count how many times a word appears in a context window, they predict the context window of a word (where the more frequent words are naturally more strongly favored to be optimized towards) \citep{baroni2014}. The other major success of word2vec was their effective preprocessing steps: dynamic context windows, removing rare words, and subsampling frequent words. These steps have shown to be where a large amount of word2vec's impressive performance comes from  \citep{lessonslearned2015}.

In this work, we consider the Skip-Gram model with Negative Sampling, which is seen as (marginally) the most effective configuration of word2vec. In the Skip-Gram model, the center word's embedding is used to predict the embeddings of the context words around it. This is in contrast to Continuous Bag-of-Words, in which all of the embeddings for the context words are averaged together and used to predict the center word's embedding. Negative Sampling is an example of ``Noise-Contrastive Estimation'' where the model trains a binary classifier to discriminate between ``true'' (word,context) pairs that occur and fake ``negative sample'' pairs that were never actually observed. A ``good'' representation is able to assign high probabilities to real (w,c) pairs while also assigning low probabilities to fictitious pairs. This is much more efficient to compute than predicting the |V|-dimensional distribution over possible words any time an update is performed.

We use two runs of word2vec in our experiments:
\begin{enumerate}
    \item \textbf{Google News}. Publicly available, pretrained word vectors trained on 100 billion tokens from the news domain.
    \item \textbf{MIMIC w2v}. Running the official word2vec tool on the corpus on 500 million tokens from MIMIC.
\end{enumerate}

\subsection{Enhancing Word Embeddings with Clinical Knowledge}

In a 2014 paper, Levy and Goldberg introduce an extension of word2vec to enable ``dependency-based word embeddings''  \citep{lg:dependency2014}. The main contribution of this work was untying the ``word'' and ``context'' vocabularies, and releasing a tool (WORD2VECF) that can train word vectors on arbitrary contexts when given a sequence of (w,c) pairs. In their work, they use structured dependency parse relationships as contexts instead of positional adjacency, and they show that this formulation learns embeddings with different structures encoded.

We use this tool to introduce similarities not in grammar-space, but in CUI-space. We generate (w,c) pairs that contain not only the original position-based contexts but also (word,CUI) pairs for every CUI that word is associated with in the UMLS MRCON database. When using this tool, we perform all necessary preprocessing steps that were originally performed by word2vec, including: using a \textit{dynamic} context window (meaning the window for a given word is chosen randomly to be between 1 and 8 tokens wide), removing rare words before generating context windows, and subsampling before generating context windows. There are 2,683,398,577 word-based (w,c) context pairs and 265,699,787 ontology-based (w,CUI) context pairs. Though the training corpus only increases in size by 11\%, a (w,CUI) pair derived from the UMLS are far more informative than a single co-occurrence (w,c) counterpart, since the CUI reflects a known medical concept, whereas many word co-occurrences are spurious relationships that reflect grammar more than semantics.

We use two runs of the WORD2VECF in our experiments in order to disentangle the benefit added by CUI contexts form any differences in implementation between word2vec and WORD2VECF:
\begin{enumerate}
    \item \textbf{MIMIC mw2v: words}. Running WORD2VECF on (w,c) pairs generated from MIMIC.
    \item \textbf{AWE-CM}. Running WORD2VECF on both (w,c) pairs generated from MIMIC and (w,CUI) pairs extracted from the UMLS.
\end{enumerate}

\section{Results}

\begin{table}
\centering 
    \caption{Spearman coefficient of correlation with various experts.}
	\begin{tabular}{|l|l|l|l|}
		\hline
                  & MayoSRS & MiniMaySRS: doctors & MiniMaySRS: coders \\ \hline
Google News       &          0.128  &         0.145  &         0.302   \\ \hline
MIMIC w2v         &  \textbf{0.398} &         0.442  & \textbf{0.572}  \\ \hline
MIMIC W2VF: words &          0.324  &         0.495  &         0.489   \\ \hline
AWE-CM            &          0.365  & \textbf{0.508} &         0.514   \\ \hline
	\end{tabular}
    \label{tab:correlations}
\end{table}

We show how well each vector space model correlates with human judgment in Table \ref{tab:correlations}. For both physician and coder judgment, non-clinical vectors perform terrible, achieving correlation coefficients that are consistently 20-30 points lower than all models trained on MIMIC. We can observe differences between word2vec and WORD2VECF by comparing rows 2 and 3. WORD2VECF is built to perform the parameter estimation of that word2vec does, so by preprocessing the corpus identically to how word2vec does, these two rows ``should'' be identical. Despite not having perfect agreement, we do see that neither method is overtly better than the other, with word2vec having higher correlation with coders and WORD2VECF correlating better with doctors. However, we \textit{can} see the benefits of incorporating CUI-space similarity into word vectors, as shown by \textit{AWE-CM} outperforming \textit{MIMIC W2VF: words} on all three judgment categories. Since these models were built from the same data with the same WORD2VECF tool, the differences between them can only be attributed to the domain knowledge incorporated by the additional (word,CUI) pairs. Additionally, we observe that the vectors that are augmented with UMLS relationships (AWE-CM, row 4) achieve the best correlation with doctors, which is more relevant for most downstream predictions using word vectors such as mortality and readmission prediction, diagnosis, information retrieval, and concept extraction.





\section{Conclusions}

In this work, we have shown how to include domain knowledge into the word2vec training process using existing tools that allow for arbitrary features as ``contexts.'' This process allows models to simulate a ``must-link'' relationship between concepts that are known to be \textit{a priori} related, even if their relationship is not captured in the limited-size training data. We demonstrate that even a simple use of the medical ontology results in across-the-board improvement above text-only vectors.

We have multiple ideas to explore for future work including: morphological notions of context (e.g. char-based features: making ``hemoglobin'' and ``hemotoxin'' vectors have more similar contexts), exploring the UMLS hierarchy beyond exact-CUI-matches, quantifying the effect of domain knowledge as a function of data size, and evaluating AWE-CM vectors on additional extrinsic evaluations. In future work, we hope to use this method to build state-of-the-art word embeddings, as compared against other publically available vectors on a variety of tasks.

\section*{Acknowledgments}
This material is based upon work supported by the National Science Foundation Graduate Research Fellowship Program under Grant No. 1122374.

\bibliographystyle{abbrvnat}
\bibliography{main.bib}

\end{document}